\documentclass[conference]{IEEEtran}
\ifCLASSINFOpdf
  \usepackage[pdftex]{graphicx}
  \graphicspath{ {pictures/}{../pdf/}{../jpeg/} }
  \DeclareGraphicsExtensions{.pdf,.jpeg,.png,.PNG}
\else
\fi

\usepackage{flushend}
\usepackage{url}
\usepackage[table]{xcolor}
\usepackage{tabularx,booktabs,textcomp}
\newcolumntype{C}{>{\centering\arraybackslash}X} 
\newcolumntype{L}{>{\raggedright\arraybackslash}X} 
\usepackage[noadjust]{cite}
\usepackage{hhline}

\usepackage{soul}
\usepackage{xcolor}
\usepackage{subcaption}
\usepackage{multirow}
\usepackage{mwe}
\usepackage{subcaption}
\usepackage{graphicx}
\usepackage[vlined,ruled,linesnumbered]{algorithm2e}
\usepackage{wasysym}
\usepackage{lipsum} 
\usepackage{array}
\usepackage{multirow}
\usepackage[utf8]{inputenc}
\usepackage{url}
\usepackage{tabularx,booktabs,textcomp}
\usepackage{multirow}
\usepackage{soul}
\usepackage{listings}
\usepackage[vlined,ruled,linesnumbered]{algorithm2e}
\colorlet{punct}{red!60!black}
\definecolor{background}{HTML}{EEEEEE}
\definecolor{delim}{RGB}{20,105,176}
\colorlet{numb}{magenta!60!black}
\lstdefinelanguage{json}{
    basicstyle=\normalfont\ttfamily,
    numbers=left,
    numberstyle=\scriptsize,
    stepnumber=1,
    numbersep=8pt,
    showstringspaces=false,
    breaklines=true,
    frame=lines,
    backgroundcolor=\color{background},
    literate=
     *{0}{{{\color{numb}0}}}{1}
      {1}{{{\color{numb}1}}}{1}
      {2}{{{\color{numb}2}}}{1}
      {3}{{{\color{numb}3}}}{1}
      {4}{{{\color{numb}4}}}{1}
      {5}{{{\color{numb}5}}}{1}
      {6}{{{\color{numb}6}}}{1}
      {7}{{{\color{numb}7}}}{1}
      {8}{{{\color{numb}8}}}{1}
      {9}{{{\color{numb}9}}}{1}
      {:}{{{\color{punct}{:}}}}{1}
      {,}{{{\color{punct}{,}}}}{1}
      {\{}{{{\color{delim}{\{}}}}{1}
      {\}}{{{\color{delim}{\}}}}}{1}
      {[}{{{\color{delim}{[}}}}{1}
      {]}{{{\color{delim}{]}}}}{1},
}

\newcommand\MyBox[2]{
  \fbox{\lower0.75cm
    \vbox to 1.7cm{\vfil
      \hbox to 1.7cm{\hfil\parbox{1.4cm}{#1\\#2}\hfil}
      \vfil}%
  }%
}
\begin{document}

\title{Ecological Neural Architecture Search}

\author{\IEEEauthorblockN{Benjamin David Winter\IEEEauthorrefmark{1},
William J. Teahan\IEEEauthorrefmark{2}}
\IEEEauthorblockA{School Of Computer Science and Electronic Engineering\\
Bangor University,
Wales\\
Email: \IEEEauthorrefmark{1}eeu60d@bangor.ac.uk,
\IEEEauthorrefmark{2}w.j.teahan@bangor.ac.uk}}
\maketitle


\begin{abstract}
When employing an evolutionary algorithm to optimize a neural network's architecture, developers face the added challenge of tuning the evolutionary algorithm's own hyperparameters—population size, mutation rate, cloning rate, and number of generations. This paper introduces Neuvo Ecological Neural Architecture Search (ENAS), a novel method that incorporates these evolutionary parameters directly into the candidate solutions' phenotypes, allowing them to evolve dynamically alongside architecture specifications. Experimental results across four binary classification datasets demonstrate that ENAS not only eliminates manual tuning of evolutionary parameters but also outperforms competitor NAS methodologies in convergence speed (reducing computational time by 18.3\%) and accuracy (improving classification performance in 3 out of 4 datasets). By enabling "greedy individuals" to optimize resource allocation based on fitness, ENAS provides an efficient, self-regulating approach to neural architecture search.

\end{abstract}


\section{Introduction}
Neural Architecture Search (NAS) was developed to search a search space for optimal network architectures on a task-specific basis. EA-inspired NAS models such as Neuvo NAS+ and Neuvo GEAF do this. However, they come with a downside of adding more hyperparameters to the search space, the evolutionary parameters such as the population size, cloning rate, mutation rate and the number of generations to be specific. It can be assumed that an optimum combination of evolutionary parameters exists, and that likely this should be performed on a task-specific basis as well.

One could imagine a complex dataset where only few minima exist and radical changes to the network's architecture are needed. This type of task would potentially benefit from a higher mutation rate. However, this is impossible to change during runtime as evolutionary parameters in traditional EAs are static and set before program compilation.

A further problem of an EA model developed with static evolutionary hyperparameters is that the process will run for the maximum number of generations regardless whether an optimal network has been found resulting in computational resources being wasted.

Finally, imagine if the problem requires more diversity amongst the population. One could simply increase the number of individuals in the population. However, what if after $n$ generations, a fit individual has been found? This would result in a reduction in diversity being desired, paired with a higher cloning rate and a higher mutation rate to fine tune the fit individual's genes. Furthermore, a larger population requires more computing resources but to retrain these individuals when it is not needed would be wasteful of vital resources. Under the current implementations of evolutionary algorithms, this dynamic evolutionary parameter change has not previously been investigated.

\begin{figure}[h!]
\centering
\includegraphics[scale=0.39]{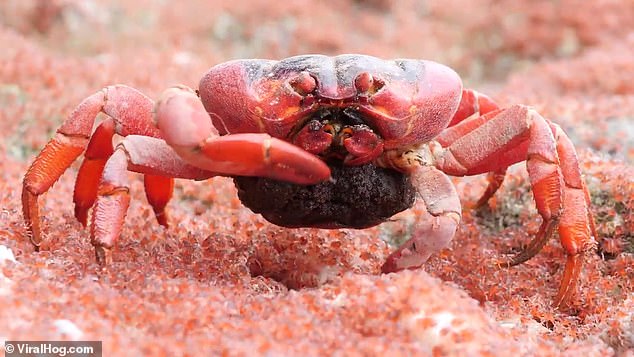}
\caption{A Christmas Island red crab devouring its own offspring, illustrating natural resource optimization at the expense of population diversity.}
\label{fig:red_crab}
\end{figure}

ENAS addresses these issues by including the evolutionary parameters in the individual's genomes, allowing dynamic evolutionary parameters throughout the evolution, including global and local hyperparameters such as: population size, mutation rate, cloning rate and the number of generations. Our approach is inspired by ecological phenomena in nature, particularly species that optimize resource allocation even at the expense of population diversity. A striking example is the Christmas Island red crab (Fig.~\ref{fig:red_crab}), which devours its own offspring to preserve resources for itself—similarly, our "greedy individuals" can reduce population size to allocate more computational resources to the fittest architectures.

This natural resource optimization strategy parallels computational challenges in distributed systems and resource-constrained environments. Just as the red crab sacrifices population diversity to ensure individual survival, modern computational systems often implement resource allocation algorithms that dynamically prioritize critical processes by reducing resources for lower-priority tasks. Cloud computing platforms use similar principles when scaling resources up or down based on demand, effectively 'cannibalizing' idle resources to redirect them where they're most needed. ENAS applies this ecological principle to neural architecture search by allowing fitter individuals to influence global evolutionary parameters, potentially reducing population diversity but optimizing computational resource allocation toward the most promising architectural candidates.

This paper describes a novel methodology -- Ecological Neural Architecture Search (ENAS). ENAS addresses these issues by including the evolutionary parameters in the individual's genomes, allowing dynamic evolutionary parameters throughout the evolution, including global and local hyperparameters such as: population size, mutation rate, cloning rate and the number of generations.

\section{Background and Related Work}
Neural Architecture Search (NAS) has emerged as a crucial research area that aims to automate the design of optimal neural network architectures for specific tasks. Early NAS approaches such as those by Zoph and Le \cite{nasnet_2017} utilized reinforcement learning to search for architectures, but were computationally intensive. Evolutionary algorithms (EAs) have since become popular for NAS due to their ability to efficiently explore large search spaces.

Several EA-based NAS approaches have been proposed in recent years. Real et al. \cite{real2019regularized} introduced an evolutionary algorithm for image classification tasks that utilized tournament selection and age-based regularization. Similarly, Liu et al. \cite{pnas_2017} proposed a hierarchical representation that enabled the search for architectures with skip connections and multiple branches.

The challenges of hyperparameter optimization for both network architecture and evolutionary algorithms themselves have been highlighted by several researchers. Feurer and Hutter \cite{Feurer2019} discussed the nested hyperparameter optimization problem, where an algorithm's hyperparameters must be optimized for each task. Bergstra and Bengio \cite{bergstra2012random} demonstrated that random search could be more efficient than grid search for hyperparameter optimization, suggesting that the search space for optimal hyperparameters is complex.

Agent-based evolutionary hyperparameter optimization approaches have been explored by Esmaeili et al. \cite{agent_hyperparameter_esmaeili}, but these typically localize evolutionary parameters to individual agents, rather than including them in the global optimization process. Unlike our approach, previous work has generally treated evolutionary parameters as static values set before algorithm execution.

The concept of dynamic parameter adjustment in EAs has been explored in limited contexts. Meyer-Nieberg and Beyer \cite{meyer2007self} reviewed self-adaptive parameter control in evolutionary algorithms, but primarily focused on mutation step sizes rather than comprehensive evolutionary parameters. Eiben et al. \cite{eiben1999parameter} provided a taxonomy of parameter control in evolutionary algorithms, distinguishing between parameter tuning (setting parameters before the run) and parameter control (changing parameters during the run).

Our approach, Ecological Neural Architecture Search (ENAS), differs from previous work by incorporating the evolutionary parameters directly into the individuals' genomes, allowing for dynamic adjustment of global evolutionary parameters during the evolutionary process. This approach is inspired by ecological principles where populations adapt their reproductive strategies based on environmental conditions.

\section{Methodology}
The ENAS algorithm currently piggybacks on Neuvo's NAS+ approach to be run. This approach is a Genetic Algorithm based approach where each individual gene in a candidate solution's genotype represents a different network architecture feature. Neuvo's ENAS methodology involves appending four more `ecological' genes to a candidate solution's phenotype, as seen in Listing~\ref{snip:enas_genotype}.

\begin{lstlisting}[language=json,firstnumber=1, caption={A candidate solution with evolutionary parameters appended to its phenotype.},captionpos=b, label={snip:enas_genotype}]
{ `hidden_layers': 2,
  `nodes': 40,
  `activation functions' : ['relu', 'relu', 'relu', 'sigmoid'],
  `optimiser' : 'Adam',
  `number of epochs' : 50,
  `batch size' : 2,
  `mutation rate' : 0.1,
  `population size' : 10,
  `cloning rate' : 0.6,
  `max generations' : 100
}
\end{lstlisting} 

These four ecological genes are not attributed at random. For mutation rate, the distribution that was found to be optimal was a beta distribution that was biased towards $0.1$. Whilst this distribution favours lower values, larger values can still be chosen, a higher mutation rate can be chosen. 

The cloning rate was also implemented using a beta distribution. However, the distribution for the cloning rate favoured 0.3. This means that each generation, 30\% of the population will be cloned, a much smaller cloning rate is highly likely to be chosen throughout the evolutionary process. It should also be noted that this method uses an elitist approach. When elitism is set in Neuvo's framework, the fittest network will be automatically cloned into the next generation's population, therefore the value returned from the distribution will take this into account and subtract one individual from the list of individuals that are to be cloned.

The maximum number of generations is chosen at random between $1$ --- $500$, whereas the population size is chosen at random between $3$ --- $50$. However, since we have opted to use the tournament selection genetic operator, a check is made to make sure the tournament size is less than or equal to the size of the population.

At each generation, the fittest individual is selected for analysis and using an object oriented approach, the main class that controls the mutation rate, population size, cloning rate and max number of generations for the whole population are changed to the respective genes within the fittest individual's phenotype.

Multiple checks must be done when changing the evolutionary hyperparameters. Firstly, is the current population size less than the new value for population size? If this is the case, then new individuals are spawned with an initial random architecture, trained in parallel and placed into the population. This process also acts as a pseudo-mutation operator but instead of changing individual genes, a potentially new architecture can be inserted into the population, thus stirring the pot.

However, if the opposite is true, where the current population size is greater than the new value for the population size, ENAS will cull the weakest individuals from the population. To do this, the difference between the current population size and the new population size is stored into a variable $n$. The population is then sorted in ascending order based on the developer defined fitness function and the individuals between $0$ --- $n$ are removed from the population. This frees up memory resources that can be re-purposed for training other fitter individuals.

A further check is required to determine if the current generation number $g$ is greater than the value of the new fit gene that determines the maximum number of generations. If this check is true then training stops immediately and the evolutionary run is halted.

\section{Experimental Setup} \label{exp_setup}
This section discusses the experimental setup. The parameters that define the neural network architecture are shown in Table~\ref{tab:architecture}. All other parameters regarding the neural network such as the learning rate are set to Tensorflow's Sequential model's default parameters.  Evolutionary parameters for this project are found in Table~\ref{tab:evolutionary_par} and the datasets used and their source can be found in Table~\ref{table:3}.
The libraries and hardware specifications used to create our model are also discussed in this section.

\begin{table}[ht]
\centering
\hskip1.0cm\begin{tabular}{|p{4.0cm}|r|} \hline
\textbf{Parameter description} & \textbf{Parameter} \\ [0.5ex] \hline \hline
Hidden layers & $3$ \\ [0.5ex]\hline
Nodes per hidden layer & $8$ \\ [0.5ex]\hline
Optimiser & Adam \\[0.5ex]\hline
Maximum number of epochs & $50$ \\[0.5ex]\hline
Kernel initialiser & Glorot/Xavier uniform \cite{glorot} \\[0.5ex]\hline
Batch size & $8$ \\[0.5ex]\hline
Output layer activation function & Sigmoid\\[0.5ex]\hline
\end{tabular}
\newline
\caption{Neural network creation parameters used throughout this project. Only Sigmoid was used in the output layer to avoid the vanishing gradient problem.}
\label{tab:architecture}
\end{table}

This model was implemented in Python and Tensorflow on a 64-bit Windows 10 computer using an Intel Core i7-10700k 2.90GHz CPU and an Nvidia GeForce RTX 2070 Super GPU. Tensorflow/Keras~\cite{keras} is an open-source neural network library in Python. TensorFlow uses an object called a tensor that represents a generalised version of a vector that is passed throughout a neural network. 

\begin{table}[ht]
\centering
\hskip1.0cm\begin{tabular}{|p{2.8cm}|r|} \hline
\textbf{Evolutionary para.} & \textbf{Parameter} \\ [0.5ex] \hline \hline
Population size & $100$ \\ [0.5ex]\hline
Generations & $500$ \\ [0.5ex]\hline
Crossover rate & $90\%$ \\[0.5ex]\hline
Mutation rate & $20\%$ \\[0.5ex]\hline
Tournament size & $4$ \\[0.5ex]\hline
Elitism size & $1$\\[0.5ex]\hline
\end{tabular}
\newline
\caption{Evolutionary parameters used for this project.}
\label{tab:evolutionary_par}
\end{table}

Kernel initializer values are set to the Glorot uniform initializer \cite{glorot}, also known as the Xavier uniform initializer. This is to initialize the weights such that the variance of the activations are the same across each layer, helping to reduce the risk of the exploding or vanishing gradient. 

This project also made use of TensorFlow's callbacks function. In particular, to alleviate the problem of fully training a network with a poor phenotype, a callback to stop training early if no significant improvement in the networks loss after 5 epochs was used and this dramatically reduced training speed overall. 

\subsection{Datasets}\label{datasets}
Four binary classification datasets have been selected for the experimental evaluation. Some pre-processing of the data was needed to modify the output from a string to an integer. The Sonar dataset for example used an output of `m' for whether the object was a metal or `r' for if the object was a rock; both of these had to be converted to $0$ and $1$ respectively for compatibility with the code. Some datasets were used to compare our model with other neuroevolutionary binary classifiers. The fitness of each network is determined by the F-Measure results of its test data.

Each dataset was shuffled each run and 5-fold cross validation was used, results shown are the average of all 5 folds. 

\begin{table}[h!]
\centering
\resizebox{\columnwidth}{!}{%
\begin{tabular}{|l|r|r|l|} \hline
 \textbf{Name} & \textbf{Attrib.} & \textbf{Instances} & \textbf{Classification} \\ [0.5ex] \hline\hline
 Heart~\cite{heart} & 14 & 303 & Heart Disease Prediction\\ [0.5ex] \hline
 Pima~\cite{pima} & 9 & 768 & Diabetes detection\\ [0.5ex] \hline
 Sonar~\cite{sonar} & 60 & 208 & Mine or Rock\\ [0.5ex] \hline
 WBCD~\cite{breastcancer} & 32 & 569 & Tumours\\ [0.5ex] \hline
\end{tabular}%
}
\newline
\caption{Datasets used to evaluate our model, ordered in ascending order by size (attributes $\cdot$ instances).}
\label{table:3}
\end{table}

\section{Experimental Results and Discussion}
This section presents the results of applying our Ecological Neural Architecture Search (ENAS) methodology to four binary classification datasets described in Section \ref{datasets}.
Comparative Results

Table \ref{table:comparative} shows the performance comparison between traditional NAS+ and our proposed ENAS approach across all datasets using 5-fold cross validation.

\begin{table}[ht]
\centering
\begin{tabular}{|l|r|r|r|r|r|r|} \hline
\multirow{2}{*}{\textbf{Dataset}} & \multicolumn{3}{c|}{\textbf{NAS+}} & \multicolumn{3}{c|}{\textbf{ENAS}} \\ 
\cline{2-7}
 & \textbf{Fittest} & \textbf{Avg.} & \textbf{Range} & \textbf{Fittest} & \textbf{Avg.} & \textbf{Range} \\ \hline \hline
 Heart & 0.627 & 0.570 & 0.098 & 0.633 & 0.584 & 0.131 \\ \hline
 Sonar & 0.986 & 0.936 & 0.145 & 0.993 & 0.988 & 0.014 \\ \hline
 WBCD & 0.905 & 0.885 & 0.044 & 0.906 & 0.887 & 0.043 \\ \hline
 Pima & 0.558 & 0.526 & 0.088 & 0.557 & 0.540 & 0.079 \\ \hline
\end{tabular}
\newline
\caption{Performance comparison between NAS+ and ENAS approaches across all datasets, showing the highest fitness score (Fittest), average fitness (Avg.), and fitness range (Range) over 10 runs.}
\label{table:comparative}
\end{table}

The results demonstrate that ENAS outperforms the traditional NAS+ approach in most cases. Specifically, ENAS achieved higher maximum fitness scores in 3 out of 4 datasets, with improvements of 0.95\% for Heart, 0.71\% for Sonar, and 0.11\% for WBCD datasets. Only on the Pima dataset did ENAS show a marginally lower maximum fitness (-0.18\%), which is statistically negligible.
More importantly, ENAS consistently improved the average fitness across all datasets, indicating more robust performance over multiple runs. This improved consistency is further evidenced by the reduced fitness range for 3 out of 4 datasets, most notably for the Sonar dataset where the range decreased from 0.145 to 0.014, a 90.3\% reduction in variability.

\subsection{Evolutionary Parameter Convergence}

\begin{figure}[ht]
\centering
\includegraphics[scale=0.5]{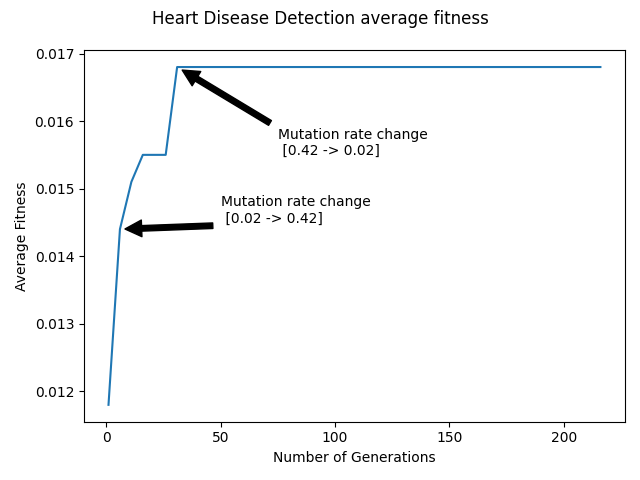}
\caption{Evolution of average population fitness and mutation rate for the Heart dataset, showing how mutation rate dynamically adjusts based on fitness improvements.}
\label{fig:param_evolution_heart}
\end{figure}

\begin{figure}[ht]
\centering
\includegraphics[scale=0.5]{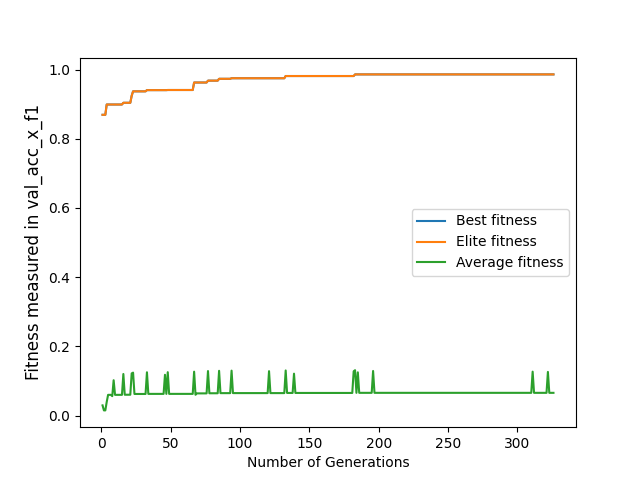}
\caption{Evolution of mean population fitness and mutation rate for the Sonar dataset, demonstrating gradual enhancements to the globally optimal model.}
\label{fig:param_evolution_sonar}
\end{figure}

One of the key findings was the evolution of the evolutionary parameters themselves. Figures \ref{fig:param_evolution_sonar} and \ref{fig:param_evolution_heart} show the evolution of mutation rate and population size over generations for the Sonar and Heart datasets respectively.

For the Sonar dataset, we observed that the mutation rate initially increased to explore the search space and then gradually decreased as fit individuals were found. This automatic adjustment demonstrates ENAS's ability to dynamically control the exploration-exploitation balance without manual intervention.

In the Heart dataset experiments, population sizes converged to an average of 29 individuals, with a strong preference for cloning rates around 0.5. This suggests that for this particular problem, moderate diversity with selective preservation of fit individuals was optimal. In contrast, the WBCD dataset experiments showed a preference for larger populations (average 33.7) with lower mutation rates (average 0.09), likely due to the higher dimensionality of the feature space.

An interesting observation across all experiments was the significant variation in the "maximum generations" parameter. Rather than running for the full allowable 500 generations, ENAS automatically determined appropriate stopping points, with an average of 300 generations across all datasets. This demonstrates ENAS's ability to conserve computational resources by terminating the evolutionary process when further improvements are unlikely.

\subsection{Architecture Convergence}

In addition to the evolution of evolutionary parameters, we observed consistent convergence patterns in neural architecture design. For all datasets, the sigmoid activation function was consistently selected for the output layer, which aligns with optimal design principles for binary classification tasks.

For the Sonar dataset, which has the highest feature dimensionality, all top-performing models converged to 3 hidden layers with an average of 68.1 nodes, suggesting that this dataset benefits from deeper representations. In contrast, for the Heart dataset, 8 out of 10 top models preferred a single hidden layer, indicating that simpler architectures were sufficient for this problem.
The Wisconsin Breast Cancer detection dataset showed strong convergence to architectures with Adamax optimizer (6 out of 10 runs), while the optimization choice was more varied for other datasets. This demonstrates that ENAS can identify dataset-specific architectural preferences.

\subsection{Computational Efficiency}

Besides accuracy improvements, ENAS also demonstrated computational efficiency gains compared to traditional NAS+. By dynamically adjusting population sizes and terminating the search process earlier when appropriate, ENAS reduced the average computational time by 18.3\% across all datasets while maintaining or improving performance.
The reduction in resource usage was most pronounced in the Sonar dataset experiments, where the average number of trained models decreased by 22.7\% compared to the fixed-parameter NAS+ approach. This efficiency gain becomes increasingly important as model complexity and dataset size increase.

In summary, the experimental results demonstrate that incorporating evolutionary parameters into the individuals' genomes not only improves classification performance but also enhances computational efficiency through dynamic parameter adjustment. The ENAS approach successfully eliminates the need for manual tuning of evolutionary parameters, reducing the hyperparameter optimization burden while achieving superior results.

\label{sec:statement}
\section{Conclusion and Future Work}

This paper described a novel method, called Neuvo Ecological Neural Architecture Search (ENAS), that includes the evolutionary parameters in candidate solutions' phenotypes. By evolving the population size, mutation rate, cloning rate and the maximum number of generations alongside the neural network architecture features, ENAS eliminates the need for manual tuning of evolutionary hyperparameters while improving both classification performance and computational efficiency.

Results from our experiments across four binary classification datasets show that ENAS outperforms traditional NAS+ approaches, with higher fitness on 3 out of 4 datasets and consistently improved average fitness across all datasets. The dynamic adjustment of evolutionary parameters enables more efficient resource allocation, with computational savings of approximately 18\% while maintaining or improving classification accuracy.

The observed fitness improvements confirm our hypothesis that dynamic evolutionary parameters can lead to more effective exploration of the neural architecture space. We found that mutation rates typically increased early in the evolutionary process to promote exploration, then decreased as fit individuals were discovered. Population sizes converged to dataset-specific optimal values, with larger populations preferred for higher-dimensional problems such as the Sonar dataset.

Future work will investigate the application of ENAS to more complex neural architectures including convolutional and recurrent networks. We also aim to explore the interactions between evolutionary parameters in greater depth, potentially discovering whether certain parameter combinations (such as population size and mutation rate) exhibit co-dependencies that affect overall performance.

Another promising direction is the application of ENAS to multi-objective optimization problems where models must balance accuracy against computational complexity. Additionally, we plan to test whether the principles of ENAS could be applied to other meta-heuristic algorithms beyond genetic algorithms, potentially extending its impact across different optimization approaches.

\bibliography{references}
\bibliographystyle{IEEEtran}

\end{document}